\documentclass[10pt,twocolumn,letterpaper]{article}

\usepackage[pagenumbers]{cvpr} 

\definecolor{cvprblue}{rgb}{0.21,0.49,0.74}
\usepackage[pagebackref,breaklinks,colorlinks,allcolors=cvprblue]{hyperref}
\usepackage{makecell}
\usepackage{multirow}

\title{RORem: Training a Robust Object Remover with Human-in-the-Loop}

\author{
    Ruibin Li\textsuperscript{1,2},
    Tao Yang\textsuperscript{3},
    Song Guo\textsuperscript{4},
    Lei Zhang\textsuperscript{1,2}\thanks{Corresponding author. This work is supported by the PolyU-OPPO Joint Innovative Research Centre.} \\
    \textsuperscript{1}The Hong Kong Polytechnic University,
    \textsuperscript{2}OPPO Research Institute,
    \textsuperscript{3}ByteDance \\
    \textsuperscript{4}The Hong Kong University of Science and Technology \\
}

\begin{document}
\maketitle

\begin{abstract}
Despite the significant advancements, existing object removal methods struggle with incomplete removal, incorrect content synthesis and blurry synthesized regions, resulting in low success rates. Such issues are mainly caused by the lack of high-quality paired training data, as well as the self-supervised training paradigm adopted in these methods, which forces the model to in-paint the masked regions, leading to ambiguity between synthesizing the masked objects and restoring the background. 
To address these issues, we propose a semi-supervised learning strategy with human-in-the-loop to create high-quality paired training data, aiming to train a \textbf{R}obust \textbf{O}bject \textbf{Rem}over (\textbf{RORem}). We first collect 60K training pairs from open-source datasets to train an initial object removal model for generating removal samples, and then utilize human feedback to select a set of high-quality object removal pairs, with which we train a discriminator to automate the following training data generation process. By iterating this process for several rounds, we finally obtain a substantial object removal dataset with over 200K pairs. Fine-tuning the pre-trained stable diffusion model with this dataset, we obtain our RORem, which demonstrates state-of-the-art object removal performance in terms of both reliability and image quality. Particularly, RORem improves the object removal success rate over previous methods by more than 18\%. The dataset, source code and trained model are available at \href{https://github.com/leeruibin/RORem}{https://github.com/leeruibin/RORem}. 

\end{abstract}

\section{Introduction}
\label{sec:intro}

\begin{figure}[t]
  \centering
   \includegraphics[width=\linewidth]{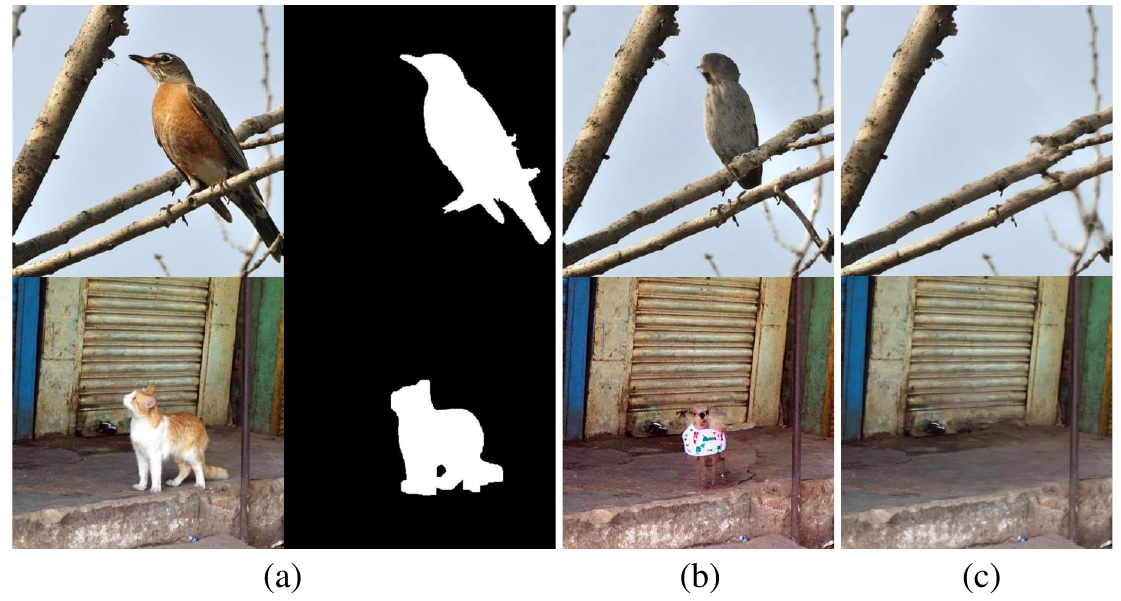}
   \caption{Given an input image and a mask (see (a)), existing object removal methods such as PowerPaint \cite{zhuang2023task} may inpaint the masked regions with other objects (see (b)), while our method can successfully remove the masked objects (see (c)).}
   \label{fig:example}
   \vspace{-5mm}
\end{figure}

Object removal aims to inpaint user-specified masked objects with realistic background, which is an important task in the fields of photography, advertising and film industry \cite{criminisi2004region,quan2024deep,zeng2020learning}. Various CNN-based \cite{pathak2016context,suvorov2022resolution,ma2022regionwise,liu2020rethinking} and transformer-based \cite{li2022mat,dong2022incremental,shamsolmoali2023transinpaint,ko2023continuously} networks have been developed, aiming to understand the image global content and thereby enhance the coherence of the inpainting process. GAN-based approaches \cite{liu2021pd,pathak2016context,wang2018perceptual,lahiri2020prior} have demonstrated their efficacy in object removal by employing adversarial loss in the training process. Recent advancements have further leveraged the generative priors from large-scale pre-trained diffusion models \cite{yildirim2023inst,guo2023shadowdiffusion,lugmayr2022repaint,xie2023smartbrush,saharia2022palette,lei2023rgbd2,winter2024objectdrop,Harmonization2024} to facilitate the inpainting of masked regions.

\begin{figure*}[t]
  \centering
   \includegraphics[width=\linewidth]{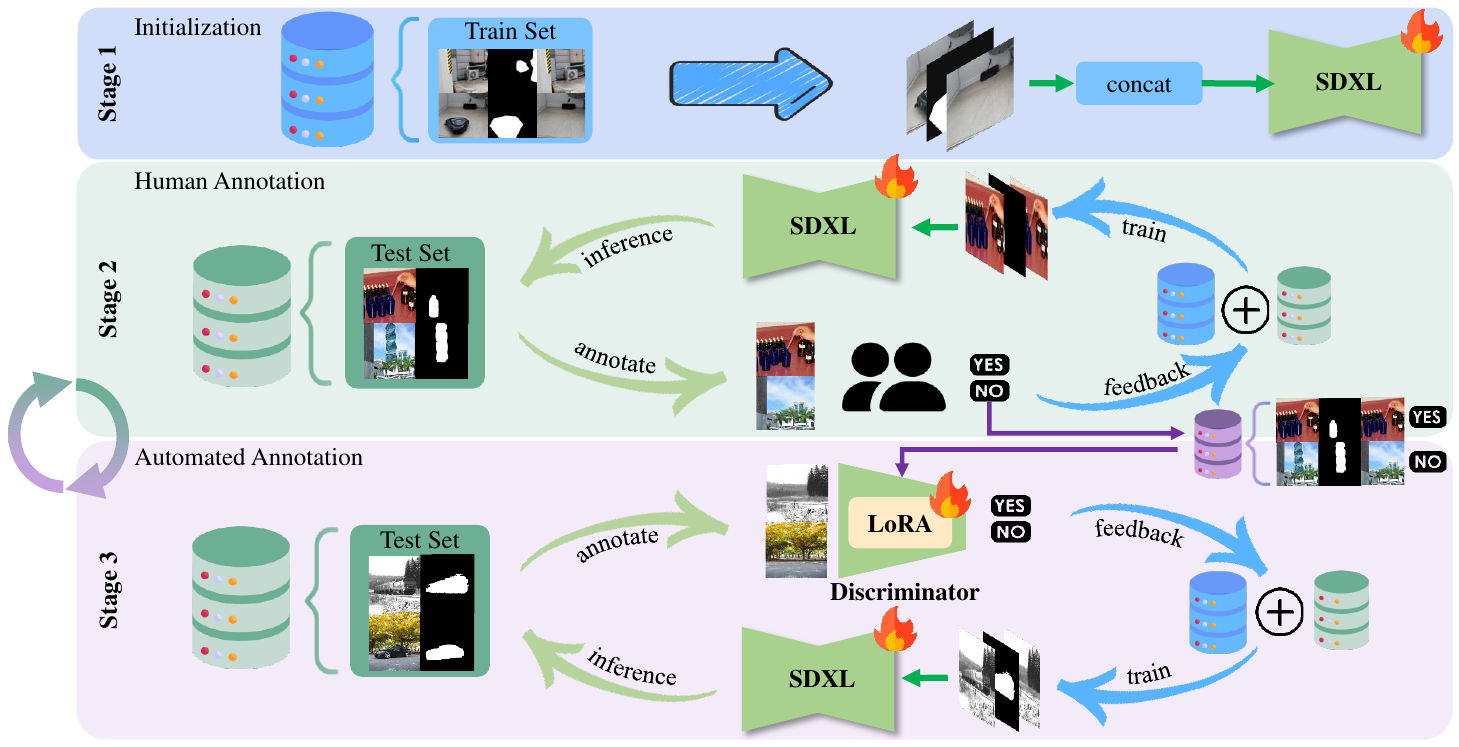}
   \caption{Overview of our training data generation and model training process. In stage 1, we gather 60K training triplets from open-source datasets to train an initial removal model. In stage 2, we apply the trained model to a test set and engage human annotators to select high-quality samples to augment the training set. In stage 3, we train a discriminator using the human feedback data, and employ it to automatically annotate high quality training samples. We iterate stages 2\&3 for several rounds, ultimately obtaining over 200K object removal training triplets as well as the trained model.}
   \label{fig:pipeline}  
   \vspace{-3mm}
\end{figure*}

Despite the commendable results, previous methods frequently encounter challenges such as incomplete removal, erroneous content synthesis and blurry synthesized regions, culminating in a low success rate. The primary reason for these shortcomings lies in the prevalent reliance on a self-supervised training paradigm using random masks \cite{winter2024objectdrop,suvorov2022resolution,podell2023sdxl}, which compels the model to inpaint the masked regions using the original content, thereby inducing ambiguities during testing. As shown in \cref{fig:example}(b), for example, when the bird or cat is masked, the training paradigm obliges the model to reconstruct the bird/cat based on the unmasked contents, whereas our objective is to remove the object and restore the background. To mitigate this ambiguity, high-quality paired training data containing images before and after the object’s presence are essential. Recent efforts have sought to construct such paired datasets, either by capturing images  \cite{sagong2022rord,winter2024objectdrop} in real-world scenarios or by synthesizing realistic data \cite{yildirim2023inst,tudosiu2024mulan}. Nonetheless, the size, diversity and quality of these datasets remain constrained, limiting the object removal performance.

To address these challenges, we propose a semi-supervised learning scheme that leverages human feedback to generate high-quality paired training data, as illustrated in \cref{fig:pipeline}. We initially collect 60K training triplets from two open-source datasets: a real video removal dataset RORD  \cite{sagong2022rord} captured by photographers, and  a synthesized dataset MULAN  \cite{tudosiu2024mulan} based on COCO \cite{lin2014microsoft} and LAION aesthetics \cite{schuhmann2022laion}. Each triplet consists of the original image, the edited image with certain items removed, and the corresponding mask. With this dataset, we train a Stable Diffusion XL (SDXL) \cite{podell2023sdxl} based inpainting model. This initial model can only achieve a success rate less than 50\% due to the limited data size and category diversity. Consequently, we introduce a human-in-the-loop approach to augment the training data. We randomly select the images and masks from the OpenImages dataset \cite{kuznetsova2020open}, and  use the initially trained model to generate the object removed samples. Then, human annotators are invited to select high-quality object removal pairs, which are then added to the training dataset.
Meanwhile, we use the human feedback data to train a discriminator that is aligned with human preference in judging high-quality object removal pairs. This discriminator enables us to automate the subsequent training data generation process. By iterating the human- and automated-annotation stages for several rounds, we obtain an object removal dataset comprising over 200K pairs across diverse categories. In addition, we compile a small high-resolution dataset for final fine-tuning to enhance the output image quality.

With our collected dataset, we fine-tune the SDXL inpainting model to obtain a \textbf{R}obust \textbf{O}bject \textbf{Rem}over, referred to as \textbf{RORem}. As shown in \cref{fig:example}(c), RORem can completely remove the targeted objects and reproduce clear background. Considering that the inference efficiency is crucial for practical usage (note that some approaches cost over 20 seconds to edit a single image), we introduce trainable LoRA layers into RORem and leverage distillation technologies  \cite{luo2023latent} to improve editing efficiency. As a result, our RORem can complete the removal process in four diffussion steps (less than 1 second). Extensive experiments demonstrate that RORem outperforms previous methods in terms of both objective metrics and subjective evaluations. Notably, for metrics such as the success rate evaluated by human subjects, RORem surpasses the second-best methods by 18\% with faster inference speed.

\section{Related Work}

\noindent\textbf{Image Inpainting.} Early image inpainting methods predominantly synthesize the training data by randomly masking regions from images. Taking the masked image as inputs, the model learns to reproduce the original masked content. Under this paradigm, early endeavors often utilize the encoder-decoder framework to accomplish the inpainting task \cite{pathak2016context,suvorov2022resolution,ma2022regionwise}, and U-Net \cite{ronneberger2015u} is widely used as the backbone  \cite{liu2018image,liu2020rethinking,yan2018shift,zeng2019learning}. In recent years, transformer-based networks have garnered increasing attention in inpainting for their intrinsic capability to complete masked patches \cite{li2022mat,dong2022incremental,shamsolmoali2023transinpaint,ko2023continuously,zhou2023propainter,cao2023zits,yu2021diverse}.
In addition, researchers have also investigated the impact of losses on inpainting performance. Beyond the conventional $L_1$ and $L_2$ losses, perceptual loss has been employed to extract high-level semantic features \cite{yang2017high,song2018contextual,li2020recurrent,liu2018image}. The GAN \cite{goodfellow2020generative} network has also been adopted for inpainting by integrating adversarial loss and trainable discriminators \cite{pathak2016context,yeh2017semantic,yu2018generative,liu2019coherent}.

Recently, diffusion models have revolutionized the field of image generation \cite{ho2020denoising,song2020denoising,rombach2022high,podell2023sdxl,esser2024scaling,blattmann2023stable}. Leveraging the powerful generative priors, recent works have adapted the pre-trained text-to-image (T2I) models for inpainting by employing the self-supervised training paradigm \cite{lugmayr2022repaint,saharia2022palette,wang2023imagen,podell2023sdxl}.
While exhibiting their efficacy across various inpainting tasks (\eg, image completion, object removal, content replacement), these methods lack reliability in large-scale tests. This limitation primarily stems from the self-supervised training paradigm, which compels the model to inpaint the masked regions utilizing the original content, which induces ambiguity during the testing phase.

\noindent\textbf{Object Removal.} Object removal can be viewed as a sub-task of image inpainting, aiming at erasing the selected objects from the given image. Therefore, many inpainting models can be directly employed for object removal tasks \cite{suvorov2022resolution,ma2022regionwise,shamsolmoali2023transinpaint,zhou2023propainter,ko2023continuously}. Among them, those stable diffusion (SD) \cite{zhuang2023task,winter2024objectdrop,yildirim2023inst,sheynin2024emu} based methods are predominantly utilized, which can can be categorized into two categories. 1) \textit{Inversion-based methods} \cite{li2024source,jia2024designedit,mokady2022null,hertz2022prompt,han2024proxedit}, which first convert the input image into a latent noise code based on inversion techniques \cite{song2020denoising,li2024source,mokady2022null}, and then modify certain intermediate features (\eg, dropping the attention feature of specific words) to yield the edited output. However, the quality of removal outputs cannot be assured in many instances, and the model efficiency is compromised due to inversion process. 2) \textit{Training-based methods}, which typically fine-tune pretrained SD models. They may utilize learnable embeddings or text prompts as auxiliary information to facilitate the object removal process \cite{zhuang2023task,brooks2023instructpix2pix,sheynin2024emu,geng2024instructdiffusion,zhang2024hive}. Some recent studies have transitioned the self-supervised training paradigm to a supervised training approach by generating removal data pairs \cite{yildirim2023inst,winter2024objectdrop,sheynin2024emu,sagong2022rord}. However, neither the datasets and models are available to the public \cite{winter2024objectdrop,sheynin2024emu,zhang2024hive}, nor the data quality and quantity are sufficient to train a reliable removal model \cite{yildirim2023inst,tudosiu2024mulan,mahfoudi2019defacto}.

\section{Proposed Method}
\label{sec:methods}
Given a source image $\mathbf{x}_s$ and a mask $\mathbf{m}$, we aim to train a generative model $G_{\theta}$ to produce an edited image $\mathbf{x}_e$ so that the unmasked region of $\mathbf{x}_e$, represented as $\mathbf{x}_e \cdot (\mathbf{1}-\mathbf{m})$, remains the same as $\mathbf{x}_s$, while the masked region, denoted as $\mathbf{x}_e \cdot \mathbf{m}$, is filled with background. $G_{\theta}$ is conventionally trained in a self-supervised manner, relying solely on training pairs of $(\mathbf{x}_s,\mathbf{m})$. However, the trained model tends to reconstruct $\mathbf{x}_s$ from the masked image $\mathbf{x}_s \cdot (\mathbf{1-m})$, resulting in ambiguity between synthesizing masked objects and restoring background.
High-quality training triplets $(\mathbf{x}_s,\mathbf{m},\mathbf{x}_e)$ can be employed to significantly enhance the removal performance since the true removal result $\mathbf{x}_e$ can circumvent the dilemma associated with the self-supervised training paradigm. Acknowledging the scarcity of high-quality triplet data, we propose a human-in-the-loop paradigm to facilitate the data generation while concurrently training the model. As illustrated in \cref{fig:pipeline}, our proposed framework is composed of an initialization stage, and human annotation stage and an automated annotation stage. The details of each stage are described in the following subsections. 

\subsection{Model Initialization}

In contrast to prior works that utilize $(\mathbf{x}_s, \mathbf{m})$ as training pairs, our approach necessitates high-quality triplet data $(\mathbf{x}_s, \mathbf{m}, \mathbf{x}_e)$ for effective model training. However, such triplet data are scarce to obtain. 
The recent work ObjectDrop \cite{winter2024objectdrop} provides 2K triplets by employing photographers to capture images before and after the object is  removed; however, the limited quantity poses challenges on model training, and neither the dataset nor the trained model is publicly available. 
Upon thorough evaluation of the existing datasets, we ultimately select two open-source datasets, RORD \cite{sagong2022rord} and Mulan \cite{tudosiu2024mulan}, to initialize our model training. 

\begin{figure}[t]
  \centering
   \includegraphics[width=\linewidth]{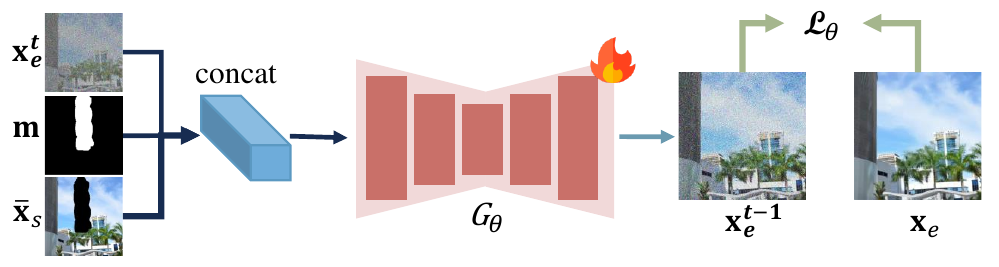}
   \caption{We finetune the pre-trained SDXL-inpaiting model with the standard diffusion training loss. We concatenate triplets data together as the model inputs. The same training paradigm is employed across all the three stages.}
   \label{fig:RORMtrain}
   \vspace{-4mm}
\end{figure}

The RORD dataset comprises about 3K short video clips captured with a fixed camera. Each video features a foreground object that moves throughout the video, with corresponding masks provided. Subsequently, a static background image of the same scene, devoid of moving objects, is captured. We extract 5 frames from each video and utilize the static image as the object removal result, yielding a total of 15K high-quality removal triplets. While the quality of this dataset is commendable, over 2.5K videos pertain to human removal cases, restricting the model's capacity to other scenes. Therefore, we incorporate a synthetic dataset, Mulan, whose images are from COCO2017 \cite{lin2014microsoft} and Laion-Aesthetics V6.5 \cite{schuhmann2022laion}. Originally designed for layered image generation, Mulan contains extracted foreground objects and inpainted background. Compared to RORD, the Mulan dataset exhibits greater diversity in categories, but its removal quality is lower.

Our training paradigm adheres to the diffusion model training pipeline, as illustrated in \cref{fig:RORMtrain}. In specific, we inject noise to the removal result $\mathbf{x}_e$ as $\mathbf{x}_e^t = \alpha_t \cdot \mathbf{x}_e + \sigma_t \cdot \epsilon$, where $t\sim [0,T]$ is the diffusion timestep, $\mathbf{\epsilon} \sim \mathcal{N}$ is the Gaussian noise, $\alpha_t,\sigma_t$ are two constants depending only on the noise scheduler and timestep $t$. 
A denoising network $G_{\theta}$, initialized by a pre-trained SDXL inpainting model \cite{podell2023sdxl}, is fine-tuned to learn the denoising process. To provide additional context regarding the background and the removal region, we concatenate the unmasked region of the source image, defined as $\bar{\mathbf{x}}_s = \mathbf{x}_s \cdot (1-\mathbf{m})$, along with the mask $\mathbf{m}$ to the noisy input $\mathbf{x}_e^t$. While ObjectDrop \cite{winter2024objectdrop} suggests that satisfactory results can be achieved by concatenating the complete source image $\mathbf{x}_s$ and the mask $\mathbf{m}$, it may result in transparent object residuals in the removal output. In contrast, we find that masking the removal region prior to concatenation can significantly enhance the robustness of object removal model. Our final loss function is defined as follows:
\begin{equation}
    \mathcal{L}_{\theta} = \operatornamewithlimits{\mathbb{E}}\limits_{t\sim [0,T],\mathbf{\epsilon} \sim \mathcal{N}}\bigg[ \| \mathbf{\epsilon} - G_{\theta}(\mathbf{x}_e^t,\bar{\mathbf{x}}_s,\mathbf{m},t) \|_2^2 \bigg].
    \label{eq:loss}
\end{equation}

\subsection{Human Annotation}
\label{sec:human_annotation}
While finetuning the SDXL-inpainting model with the initial dataset can improve the removal robustness, the success rate is hard to exceed 50\% due to the limited number and quality of training samples. We implement a human-in-the-loop process to further enhance the training dataset and our model.
Specifically, we randomly select images from the OpenImages dataset \cite{kuznetsova2020open} to construct a test set, which contains 10K pairs. Each pair consists of a source image, denoted by $\mathbf{x}_s^i$, and a mask, denoted by $\mathbf{m}^i$.
During the selection process, we exclude certain terms (\eg, clothes, body) to prevent from acquiring erroneous knowledge. Additionally, if the number of instances for a particular class exceeds 500, we will stop the sampling of this class. By applying the initial removal model to this test set, we obtain 10K removal results, denoted by ${\mathbf{x}_e^i}$, which encompass both high-quality and low-quality removal cases. 

To filter out low-quality removal cases, we engage 10 human annotators to evaluate the removal images. For each case, the annotators are provided with the source image, the mask image, and the edited image, and they are asked to assign the removal result by a label $y^i$, whose value is either ``yes" (\ie, high quality) or ``no" (\ie, low quality). This process enables us to compile a quadruple set ${(\mathbf{x}_e^i, \mathbf{x}_s^i, \mathbf{m}^i, y^i)}$. During the annotation process, those cases with incomplete removals, blurry removal regions, and incorrect inpainting contents are all classified as failure cases. With human feedback, we can expand the training dataset with  high-quality removal samples and retrain our model $G_{\theta}$. Specifically, we collect 4182, 7008 and 6133 valid removal samples, respectively, in three rounds of human annotations, as shown in \cref{tab:train_data}.

\subsection{Automated Annotation}
\label{section:automate}

\begin{figure}[t]
  \centering
   \includegraphics[width=\linewidth]{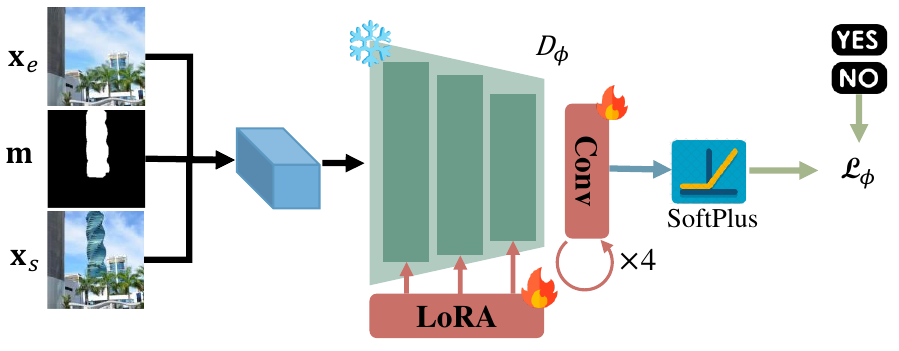}
   \caption{Training of the discriminator for automated data annotation. We use the down and middle blocks of SDXL-inpainting model as the base model, introduce trainable LoRA layers into it, and add several convolutional layers after them. Human feedback data are utilized to train the LoRA and convolutional layers.}
   \label{fig:discriminator}
   \vspace{-3mm}
\end{figure}

While human annotation can output high-quality removal samples, it is very costly and time-consuming. To collect data more cost-effectively, we propose to use the quadruple set ${(\mathbf{x}_e^i, \mathbf{x}_s^i, \mathbf{m}^i, y^i)}$ collected in the human annotation process to train a discriminator, denoted by $D_{\phi}$, and use it to perform automated annotation.
The architecture and training framework of discriminator $D_{\phi}$ are depicted in \cref{fig:discriminator}. We leverage the down and middle blocks of pre-trained SDXL-inpainting model \cite{podell2023sdxl} as the backbone, and introduce trainable LoRA \cite{hu2021lora} layers (with rank 4) to fine-tune it. Additionally, we introduce several convolutional layers to transform the middle block output into a confidence score ranging from 0 to 1. The training loss of $D_{\phi}$ is:
\begin{equation}
    \mathcal{L}_{\phi} = \frac{1}{N} \sum_{i=1}^{N}\| D_{\phi}(\mathbf{x}_e^i,\mathbf{x}_s^i,\mathbf{m}^i) - y^i \|_2^2.
    \label{eq:discriminator}
\end{equation}
The discriminator takes the object removed image $\mathbf{x}_e$, the source image $\mathbf{x}_s$ and the mask image $\mathbf{m}$ as input, and uses the human annotated label as the supervision output. More details on the training of $D_{\phi}$ can be found in Sec. \ref{sec:train_discriminator}.

Once $D_{\phi}$ is trained, we employ the same sampling principle as in human annotation to collect another test set from the Openimage dataset, and apply the removal model $G_{\theta}$ retrained in the human annotation stage to this test set. The discriminator $D_{\phi}$ is used to label the removal results. To ensure the quality of automatic labeling, only those removal samples whose confidence scores are higher than 0.9 are selected as the successful cases and added to the training set.  

\begin{table*}[t]
  \centering
  \begin{tabular}{@{}lcccccc@{}}
    \toprule
    Round & Datasets & \makecell[c]{No. of\\Test Images} & \makecell[c]{No. of\\Selected Pairs} & Total Train Size & Success Rate & PSNR\\
    \midrule
    Base Model & --- & --- & --- & --- & 7.6 & 25.72\\
    \midrule
    Initialization & RORD\&Mulan & 61,565 & 61,565 & 61,565 & 38.6 & 28.41\\
    Human (Round 1) & OpenImage & 10,000 & 4,182 & 65,747 & 47.8 & 28.63\\
    Automation (Round 1) & OpenImage & 30,000 & 20,634 & 86,381 & 55.6 & 28.60\\
    Human (Round 2)& OpenImage & 10,000 & 7,008 & 93,389 & 61.4& 28.70\\
    Automation (Round 2)& OpenImage & 80,000 & 51,099 & 144,488 & 67.2 & 28.75\\
    Human (Round 3)& OpenImage & 10,000 & 6,133 & 150,621 & 71.8 & 28.77\\
    Automation (Round 3)& OpenImage & 95,204 & 49,313 & 199,934 & 75.4 & 28.78\\
    \midrule
    Final Stage & DIV2K\&Flicker2K & --- & 1,200 & 201,134 & 76.2 & 31.10\\
    \bottomrule
  \end{tabular}
\vspace{-2mm}
  \caption{The details of our constructed dataset throughout the several rounds of annotations. We employ SDXL-inpainting as the initial object removal model and fine-tune it during our dataset construction process.}
  \label{tab:train_data}
\end{table*}

We iterate the human and automated annotation stages for 3 rounds to increase the size and diversity of our training set. The numbers of test images and selected pairs in each round are shown in \cref{tab:train_data}. After each round of human annotation, we fine-tune the discriminator to improve its discrimination performance. 
After the final stage of automated annotation and re-training, we compile a small dataset of 1200 high-quality removal pairs. The images of this dataset are selected from the DIV2K \cite{Agustsson_2017_CVPR_Workshops} and Flicker2K \cite{timofte2017ntire} datasets, which have 2K resolution. SAM \cite{kirillov2023segment} is used to generate the masks. We utilize this dataset to conduct the final 20K fine-tuning steps, aiming at enhancing the overall image quality of removal results. As shown in \cref{tab:train_data}, while further improving a little the removal success rate, the fidelity metric PSNR is significantly improved.

\subsection{Model Distillation}

\begin{figure}[t]
  \centering
   \includegraphics[width=\linewidth]{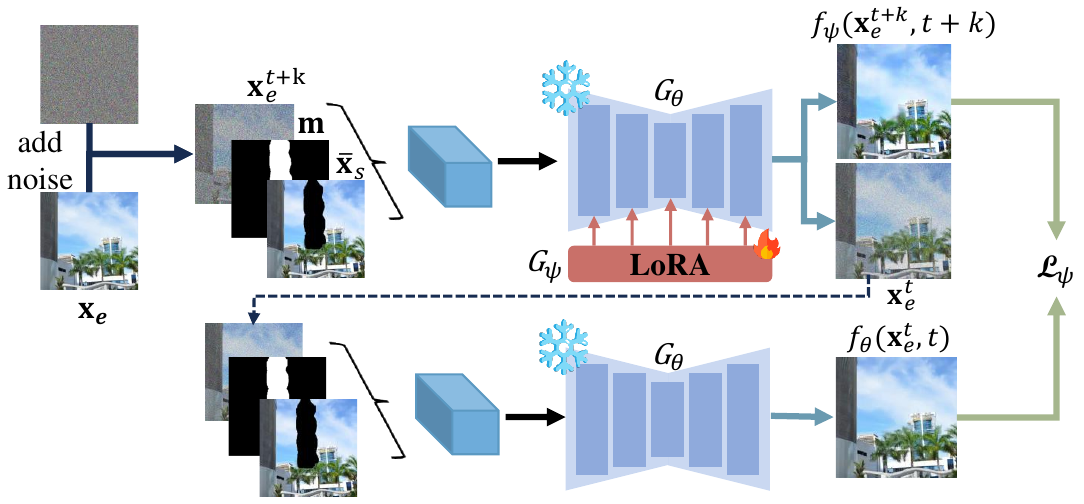}
   \caption{Efficient model distillation. We integrate trainable LoRA layers into the trained RORem model, and fine-tune it by adapting the pipeline of latent-consistency-model (LCM) under the guidance of original RORem. The distilled model can perform high-quality object removal in four diffusion steps.}
   \label{fig:distillation}
   \vspace{-3mm}
\end{figure}

While our RORem can produce promising removal outcomes, it takes tens of diffusion steps to complete the removal process, which incurs a running time of over 4 seconds per image on an A100 GPU (50 steps). To improve time efficiency, we propose to introduce trainable LoRA layers with a rank of 64 into RORem, and adopt distillation techniques \cite{luo2023latent} to distill a four-step RORem. The distillation process is illustrated in \cref{fig:distillation}. 

For the convenience of expression, we denote by $G_{\theta}(\mathbf{x}_e^t,t)$ the predicted noise $G_{\theta}(\mathbf{x}_e^t,\bar{\mathbf{x}}_s,\mathbf{m},t)$. To perform distillation, we first define a function $f(\mathbf{x}_e^t,t)$, which aims to estimate the clear image from $G_{\theta}(\mathbf{x}_e^t,t)$. The function $f(\mathbf{x}_e^t,t)$ is defined as follows:
\begin{equation}
    f_{\theta}(\mathbf{x}_e^t,t) = c_{skip}(t) \mathbf{x}_e^t + c_{out}(t) \frac{\mathbf{x}_e^t - \sigma_t G_{\theta}(\mathbf{x}_e^t,t) }{\alpha_t},
    \label{eq:mapping}
\end{equation}
where scalars $c_{skip}(t)$ and $c_{out}(t)$ depend solely on the noise scheduler and the timestep $t$ \cite{song2020denoising,ho2020denoising}. As illustrated in \cref{fig:distillation}, we denote by $f_{\theta}$ and $f_{\psi}$ the above functions associated with the original RORem $G_{\theta}$ and the fine-tuned RORem with LoRA layers $G_{\psi}$, respectively. We expect that the output image of $f_{\psi}$ should be as close to that of $f_{\theta}$ as possible, which can be achieved by employing the following distillation loss function:
\begin{equation}
    \mathcal{L}_{\psi} = \operatornamewithlimits{\mathbb{E}}\limits_{t\sim [0,T]} \| f_{\psi}(\mathbf{x}_e^{t+k},t+k) - f_{\theta}(\hat{\mathbf{x}}_e^t,t) \|_2^2,
    \label{eq:distillation}
\end{equation}
where $\hat{\mathbf{x}}_e^t$ is estimated by denoising $\mathbf{x}_e^{t+k}$ for $k$ steps based on DDIM sampling \cite{song2020denoising}, represented as $\hat{\mathbf{x}}_e^{t}=DDIM_{\theta}(\mathbf{x}_e^{t+k},\bar{\mathbf{x}}_s,\mathbf{m},t+k)$. In our experiments, we set $k=20$, which is consistent with LCM \cite{luo2023latent}. All parameters $\theta$ in our RORem are fixed throughout the distillation process. Unlike the original LCM, we set the text condition as null $\varnothing$ and the classifier-free guidance scale to 1, as text condition input is unnecessary in the our task of object removal process. This modification not only reduces memory requirements but also enhances the efficiency during both training and inference. We fine-tune the LoRA parameters $\psi$ for 30,000 steps utilizing our constructed removal dataset, enabling us to complete the removal process in four steps with an average runtime of 0.50 second on an A100 GPU.

\section{Experiment}

\subsection{Experiment Setting}
\label{sec:4-1}
\begin{figure}[t]
  \centering
   \includegraphics[width=0.85\linewidth]{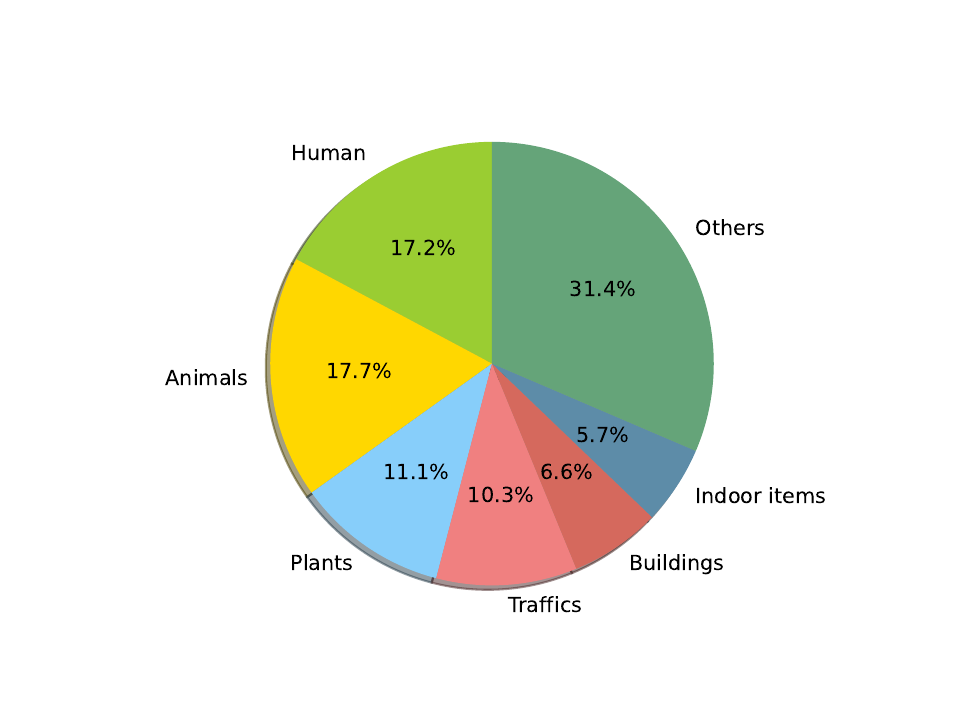}
   \caption{The category distribution of our constructed dataset.}
   \label{fig:distribution}
   \vspace{-4mm}
\end{figure}

\noindent\textbf{Training and Testing Datsets.}
As described in Sec. \ref{sec:methods}, we utilize RORD \cite{sagong2022rord} and Mulan \cite{tudosiu2024mulan} as our initial training datasets. Consequently, we sample images from the OpenImage dataset \cite{kuznetsova2020open}, filtering out images by specific keywords (\eg, clothes, human body) to mitigate unreasonable removal cases and excluding instances with excessively small ($<3\%$) or large ($>70\%$) mask regions. For each sampled image from the OpenImage dataset, we resize its shortest side to 512 pixels and center-crop the image to a resolution of $512 \times 512$ for training. We employ both human and automatic annotations to augment our training dataset, ultimately yielding a total of 201,134 removal pairs, as summarized in \cref{tab:train_data}. The  category distribution  of our final training dataset is illustrated in \cref{fig:distribution}.

We evaluate RORem alongside other competing methods using two test sets, which have the same image scenes but under two resolutions: $512 \times 512$ and $1024 \times 1024$. Both test sets have 500 pairs of original images and their corresponding masks. The test images are also sampled from the OpenImage dataset and preprocessed using the same procedures as we employed for the training data. Since methods like PPT perform poorly with fine-grained masks, we dilate the mask with Open Computer Vision Library (OpenCV2) \cite{opencv} with dilation kernel sizes as 50 and 100 for 512 and 1024 resolutions, respectively. 

\noindent\textbf{Model Training.} We train RORem using the AdamW optimizer \cite{loshchilov2017decoupled} with a learning rate of 5e-5. In each round of training, we perform 50K optimization steps with a total training batch size of 192 across 16 NVIDIA A100 GPUs. The SDXL-inpainting model \cite{podell2023sdxl} is employed as the initial model for fine-tuning. During the training process, we set the text prompt to null, as it is unnecessary for our RORem. 

\noindent\textbf{Compared Methods.}
We compare RORem with state-of-the-art object removal methods, including Lama \cite{suvorov2022resolution}, SDXL-inpainting (SDXL-INP) \cite{podell2023sdxl}, Inst-inpainting (INST) \cite{yildirim2023inst}, PowerPaint (PPT) \cite{zhuang2023task}, Instructpix2pix (IP2P) \cite{brooks2023instructpix2pix}, CLIPAway \cite{ekin2024clipaway}, and DesignEdit \cite{jia2024designedit}. Except for Lama, all the other methods leverage the pre-trained SD model. DesignEdit is a training-free method, built upon the noise inversion techniques and the inference framework of SD. For SDXL-inpainting, we employ LLava-1.6 \cite{liu2024visual} to generate captions for the background, and use them as text prompts for image completion. Note that we do not compare RORem with the ObjectDrop \cite{winter2024objectdrop} and EmuEdit \cite{sheynin2024emu}, as their source codes or models are not publicly available.

\begin{figure*}[t]
  \centering
   \includegraphics[width=\linewidth]{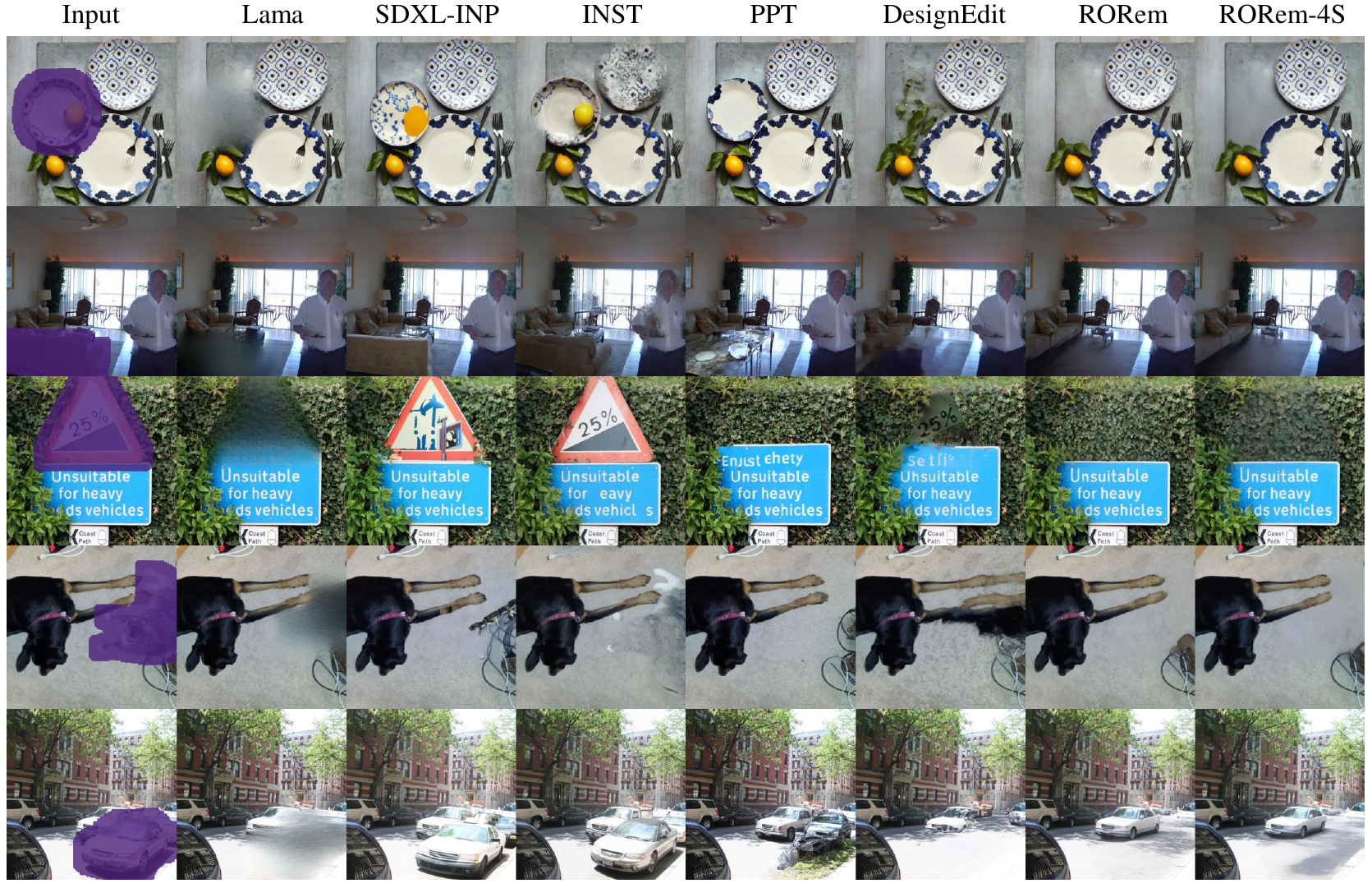}
\vspace{-3mm}
   \caption{Visual comparison of the object removal results by RORem and other methods on $1024 \times 1024$ resolution images. One can see that there can be incomplete removal regions, blurry synthesis output, and incorrect synthesis contents in previous methods, while RORem demonstrates robust removal performance. Due to limited space, we put the visual results of IP2P, CLIPAway in the \textbf{\cref{sec:SP4}}.}
   \label{fig:qualitative}  
   \vspace{-5mm}
\end{figure*}

\noindent\textbf{Evaluation Metrics.} Following prior works \cite{zhuang2023task,winter2024objectdrop,yildirim2023inst}, we employ the classical fidelity metric PSNR, alongside perceptual metrics DINO \cite{oquab2023dinov2}, CLIP \cite{radford2021learning}, and LPIPS \cite{zhang2018unreasonable}, to comprehensively assess the competing methods. 
Considering that these metrics may not be able to accurately reflect the practical object removal performance, we conduct a user study by inviting five volunteers. The volunteers are presented with the object removal outcomes generated by each method together with the original image and the mask. They are asked to determine whether the model output is a success or a failure, based on factors such as whether the object is completely removed and the quality of the object removed image. The success rate of each method is computed by averaging the results across all volunteers. The interface and more details of the user study can be found in the \textbf{\cref{sec:SP1}}. In addition, we also utilize our trained discriminator $D_{\phi}$, which aims to approximate human judgment on the success or failure of object removal output, as another metric in the experiment.

\subsection{Object Removal Results}
\label{sec:4-2}
As shown in \cref{tab:train_data}, the size of our training dataset increases from the initial 60K pairs to the final 200K pairs. With more high quality training pairs, the removal success rate of our RORem model escalates from 7.6\% to 76.2\% on our test set. The visual examples of object removal results at each training round can be found in the \textbf{\cref{sec:SP3}}. One can see that with the expansion of training data, RORem gradually improves its object removal robustness and the removal quality. In the following, we compare our RORem model with its competitors quantitatively and qualitatively.

\noindent\textbf{Quantitative Comparisons.} The quantitative comparisons among the competing methods are presented in \cref{tab:quantitative}. We have the following observations. First, RORem demonstrates substantial improvements over previous methods in terms of success rate via user study. In particular, RORem's success rate is about 18\% higher than the second best methods on both resolutions $512 \times 512$ and $1024 \times 1024$. This validates the exceptional robustness of RORem. Lama works well when the image size is $512 \times 512$, but its limited generative capacity makes it exhibit poor performance on higher resolution images. SDXL-based methods like SDXL-INP and DesignEdit perform better on $1024 \times 1024$ resolution images but they lag much behind our RORem. Overall, the competing methods may achieve good results on some cases, but their robustness is rather limited. Similar trend can be observed when evaluating the success rates using our trained discriminator $D_{\phi}$. Specifically, based on $D_{\phi}$, RORem's success rate is approximately 18\% and 15\% higher than the second-best methods at resolutions of $512 \times 512$ and $1024 \times 1024$, respectively. Furthermore, we can see that the success rates estimated by $D_{\phi}$ closely align with human annotation in the test set (the deviation is less than 3\% in most cases). This indicates that our trained $ D_{\phi} $ effectively mirrors human preferences.

Second, RORem achieves the best PSNR metric among the diffusion-based methods, and it is only lower than Lama. This is because the diffusion-based methods utilize VAEs to compress images to the latent space, leading to some loss of details in the unmasked regions. Since Lama does not rely on diffusion models, it achieves high PSNR on unmasked regions. However, its inpainting quality on the masked region is not as good as RORem. 
Third, RORem achieves state-of-the-art performance in perceptual metrics, including LPIPS, DINO and CLIP score on both resolutions, producing removal results that are more consistent with human perception. Finally, our distilled model with four diffusion steps, denoted by RORem-4S, achieves close performance to RORem but with a significant reduction in inference time from 4.03s and 4.44s to 0.50s and 0.83s per image (a reduction of 88\% and 81\%) on the two resolutions. Though there are some deterioration in PSNR and LPIPS, there is only a slight reduction in the success rate (1.4\% and 2.8\%). Overall, RORem-4S achieves the second best results in success rate, DINO, CLIP and inference time.

\begin{table*}[t]
  \centering
  \begin{tabular}{@{}c|lccccccc@{}}
    \toprule
    \makecell{Image \\ Size} & Method & 
    \makecell{Success Rate \\ (Human)$\uparrow$} & \makecell{Success Rate \\ ($D_{\phi}$)$\uparrow$} & PSNR$\uparrow$ & LPIPS$\downarrow$ & DINO$\uparrow$ & CLIP$\uparrow$ &  Time(s)\\
    \midrule
    \multirow{8}*{\rotatebox{90}{$512 \times 512$}} & Lama (WACV 2022 \cite{suvorov2022resolution}) & 55.4 & 48.6 & \textbf{33.06} & \underline{2.99}  & \textbf{0.77} & 22.77 & \textbf{0.15}\\
    \cline{2-9}
    ~ & SDXL-INP (ICLR 2024 \cite{podell2023sdxl}) & 15.8 & 16.0 & 26.03 & 4.72  & \underline{0.76} & 22.23 & 4.52\\
    ~ & INST (ArXiv 2023 \cite{yildirim2023inst}) & 3.0 & 3.6 & 23.75 & 10.36 & 0.74 & 20.80 & 5.68\\
    ~ & PPT (ECCV 2024 \cite{zhuang2023task}) & 55.8 & 56.8 & 28.41 & 6.06  & 0.75 & 22.36 & 1.98\\
    ~ & IP2P (CVPR 2023 \cite{brooks2023instructpix2pix}) & 10.0 & 7.2 & 19.81 & 19.95 & 0.72 & 13.82 & 3.62\\
    ~ & CLIPAway (NIPS 2024 \cite{ekin2024clipaway}) & 32.4 & 35.4 & 28.69 & 3.55 & 0.75 & 22.91 & 4.95\\
    ~ & DesignEdit (ArXiv 2024 \cite{jia2024designedit}) & 29.2 & 25.4 & 26.79 & 8.16  & 0.68 & 22.01 & 12.88\\
    \cline{2-9}
    ~ & RORem & \textbf{76.2} & \textbf{74.6} & \underline{31.10} & \textbf{2.86}  & \textbf{0.77} & \textbf{23.28} & 4.03\\
    ~ & RORem-4S & \underline{74.8} & \underline{71.0} & 30.08 & 3.47 & \textbf{0.77} & \underline{23.24} & \underline{0.50}\\
    \midrule
    \midrule
    \multirow{8}*{\rotatebox{90}{$1024 \times 1024$}} & Lama (WACV 2022 \cite{suvorov2022resolution}) & 18.2 & 13.8 & \textbf{36.36} & 2.68 & 0.75 & 22.24 & \textbf{0.21} \\
    \cline{2-9}
    ~ & SDXL-INP (ICLR 2024 \cite{podell2023sdxl}) & 20.4 & 18.6 & 32.28 & \underline{1.59} & \underline{0.77} & 22.06 & 5.57\\
    ~ & INST (ArXiv 2023 \cite{yildirim2023inst}) & 3.2 & 3.6 & 24.83 & 9.42 & \underline{0.77} & 20.65 & 6.75\\
    ~ & PPT (ECCV 2024 \cite{zhuang2023task})& 46.8 & 54.6 & 33.34 & 2.40 & \underline{0.77} & 22.62 & 5.61\\
    ~ & IP2P (CVPR 2023 \cite{brooks2023instructpix2pix})& 6.6 & 4.6 & 22.89 & 24.90 & 0.76 & 20.75 & 10.46\\
    ~ & CLIPAway (NIPS 2024 \cite{ekin2024clipaway})& 23.8 & 29.2 & 33.04 & 1.86 & 0.75 & 22.42 & 41.42\\
    ~ & DesignEdit (ArXiv 2024 \cite{jia2024designedit}) & 52.4 & 56.8 & 32.98 & 2.73 & \underline{0.77} & 22.97 & 23.11\\
    \cline{2-9}
    ~ & RORem & \textbf{70.2} & \textbf{71.6} & \underline{36.05} & \textbf{1.44} & \textbf{0.78} & \textbf{23.14} & 4.44 \\
    ~ & RORem-4S & \underline{67.4} & \underline{68.2} & 35.02 & 1.84 & \underline{0.77} & \underline{23.03} & \underline{0.83} \\
    \bottomrule
  \end{tabular}
\vspace{-3mm}
  \caption{The quantitative comparison of different object removal methods under two image resolutions. The best and second best results of each metric are highlighted in bold and underscore, respectively. RORem-4S means our distilled RORem model with 4 diffusion steps.}
  \label{tab:quantitative}
  \vspace{-4mm}
\end{table*}

\noindent\textbf{Qualitative Comparisons.} The qualitative comparison on images of resolution $1024 \times 1024$ are illustrated in \cref{fig:qualitative}. We can see that Lama generates blurry synthesis output in most cases and exhibits poor generation quality. SDXL-INP and INST fail in most cases. The masked regions are partially removed (see images plate in row 1 and cat in row 4) or not removed (see images sign and car).  PPT sometimes fills the masked regions with incorrect contents (see images sofa and car). Especially, the sign is inpainted with nonexistent words in row 3. While DesignEdit successfully removes the car in row 5, it suffers from visual artifacts in images sofa and sign. Meanwhile, the surrounding items can impose negative impact on the synthesis output, leading to wrong filling contents in images plate and dog. 
In contrast, RORem successfully removes the masked regions in all cases. Furthermore, our distilled RORem-4S model also works well in just four diffusion steps. Due to limited space, we put the visual results on images of $512 \times 512$ resolution in the \textbf{\cref{sec:SP5}}. 

\begin{table}[t]
  \centering
  \begin{tabular}{@{}lcccc@{}}
    \toprule
    Method & Precision$\uparrow$ & Recall$\uparrow$ & F1$\uparrow$ & Acc.$\uparrow$\\
    \midrule
    $D_{\phi}$ (Round 1) & 0.833 & 0.528 & 0.646 & 0.685\\
    $D_{\phi}$ (Round 2) & 0.901 & 0.565 & 0.695 & 0.706\\
    $D_{\phi}$ (Round 3) & \textbf{0.987} & 0.466 & 0.633 & 0.680\\
    \midrule
    $D_{\phi}$ (+Synthesized) & 0.621 & 0.921 & 0.742 & 0.669 \\
    $D_{\phi}$ (+Annotated) & 0.740 & 0.890 & 0.808 & 0.782 \\
    $D_{\phi}$ (+All) & 0.821 & 0.840 & 0.830 & \textbf{0.823} \\
    \bottomrule
  \end{tabular}
\vspace{-3mm}
  \caption{Performance of discriminator in different rounds. Acc. means Accuracy.}
  \label{tab:discriminator}
  \vspace{-3mm}
\end{table}

\subsection{Training and Evaluation of Discriminator}
\label{sec:train_discriminator}

In our dataset construction process, we train a  discriminator $ D_{\phi} $ to automate the training sample selection process. Therefore, it is necessary to evaluate whether the discrimination capability of $ D_{\phi} $ is good enough to align with human preference. We use our human labeled test set to assess $ D_{\phi} $ in this section, and the results are shown in \cref{tab:discriminator}. (For the definitions of precision, recall, F1 and accuracy, please refer to the \textbf{\cref{sec:SP2}}.)

\begin{table*}[t]
  \centering
  \begin{tabular}{@{}ccccccccc@{}}
    \toprule
     & \multicolumn{2}{c}{Annotated Data} & \multicolumn{6}{c}{Synthesized Data}\\
     \cmidrule(r){2-3} \cmidrule(l){4-9} 
     & RORem & Baselines & Blur & Noise & Downsample & Mixed & No-change & RORD \\
    \midrule
    Positive & 17,322 & 785 & 0 & 0 & 0 & 0 &0 & 18,859  \\
    Negative & 12,678 & 3,415 & 3,000 & 3,000 & 3,000 & 3,000 & 3,000 & 0  \\
    \midrule
    Total & 30,000 & 4,200 & 3,000 & 3,000 & 3,000 & 3,000 & 3,000 & 18,859  \\
    \bottomrule
  \end{tabular}
\vspace{-1mm}
  \caption{The details of our constructed dataset for training the final discriminator. `Baseline' means the seven competing methods used in the experiments. `Mixed' refers to the combination of Blur, Noise and Downsample degradations. `No-change' indicates the use of the source image directly as the editing result.}
  \label{tab:enhanceData}
  \vspace{-3mm}
\end{table*}

Since high-quality removal samples are crucial for model training, we choose the discriminator checkpoints with the highest precision and set a high threshold as 0.9. This can ensure that the selected removal pairs are of high-quality in each round of automation annotation. The performance of  $ D_{\phi} $ trained in the 3 rounds of automated annotation are presented in the top three rows of \cref{tab:discriminator}. In the initial stage, with a human feedback training dataset comprising 10K samples, the precision of $D_{\phi}$ is 0.833. As the training dataset expands, the discriminator's precision improves significantly, exceeding 0.98 in the last round. While the score on other metrics may not as good as the precision, this ensures that only the very high quality samples will be included in the training data expansion process. 

After expanding the dataset, how to ensure the accuracy becomes crucial for reliable performance evaluation. We observe that while $ D_{\phi} $ aligns well with human preferences when evaluating our RORem, it exhibits bias in assessing other methods because $ D_{\phi} $ is exposed only to the failure cases of RORem during training. To make $ D_{\phi} $ a good assessor for more competing methods, we expand the training data of it using several strategies, as detailed in \cref{tab:enhanceData}.
First, in addition to the human annotated 17,322 positive and 12,678 negative samples of RORem, we sample 600 examples from our training dataset and edit them using the seven competing methods. The edited results are manually annotated, leading to 785 positive and 3,415 negative samples. Second, we apply various degradation (blur, noise, dowmsample and the mixture of them) and `no-change' to the masked regions of RORem editing outputs, generating 15,000 negative samples. Finally, we consider all the 18,859 pairs in the RORD dataset as positive samples.

With this enriched dataset, we fine-tune the discriminator $ D_{\phi} $ for additional training rounds. We experiment with three settings: using synthesized data along with RORem annotated data, using baseline annotated data along with RORem annotated data, and using all the synthesized and annotated data as the training set. The performance of $ D_{\phi} $ is presented in the three bottom rows of \cref{tab:discriminator}.
Our results indicate that enriching the training data with synthesized samples improves recall, allowing $ D_{\phi} $ to better recognize positive samples. 
By employing both synthesized and annotated data, we can further improve accuracy. Finally, by designating removal pairs with predicted scores lower than 0.35 as negative samples, the discriminator achieves an accuracy of 0.823.
We then utilize this refined $ D_{\phi} $ to evaluate the success rates of different methods, with results presented in \cref{tab:quantitative}. The results demonstrate that $ D_{\phi} $ effectively aligns with human preferences.

\section{Conclusion}
\label{sec:conclusion}
We proposed RORem, a robust object removal model with human-in-the-loop during training. To assemble a large-scale, high-quality, and diverse removal dataset, we introduced a semi-supervised learning scheme that leverages both human and automatic annotations, ultimately building a dataset with 200K high-quality object removal pairs. Utilizing this dataset, we fine-tuned an SDXL-based inpainting model into a reliable removal model, which was further distilled into four diffusion steps to  facilitate inference speed. Experimental results demonstrated the outstanding object removal performance of RORem. In specific, it achieved about 18\% higher success rate than previous methods on two different resolutions.

Despite its clear advantages, RORem has certain limitations. First, due to the inherent problems of VAEs in image detail compression, the image quality of unmasked regions may suffer some degradation. Second, although RORem steers pretrained model toward the specific task of object removal and background filling, it still encounters challenges in achieving satisfactory results when the background contains human fingers and small faces (see the \textbf{\cref{sec:SP6}} for visual examples), which is also a known problem for many generative diffusion models. We believe that one potential solution is to leverage more advanced foundation models, such as SD3 or Flux and we leave this as our future work.

{
    \small
    \bibliographystyle{ieeenat_fullname}
    \bibliography{main}
}

\clearpage
\appendix
\setcounter{page}{1}
\onecolumn

\begin{center}
    \Large
    \textbf{\thetitle} \\ 
    \vspace{0.5em} 
    Supplementary Material \\
    \vspace{1.0em} 
\end{center}

In this supplementary file, we provide the following materials:

\begin{itemize}[leftmargin=2em]
    \item The interface and more details of the user study (referring to Sec. \ref{sec:4-1} in the main paper);
    \item Details of the evaluation metrics for the discriminator (referring to Sec. \ref{sec:4-1} and Sec. \ref{sec:train_discriminator} in the main paper);
    \item Visual examples of object removal results at each training round (referring to Sec \ref{sec:4-2} in the main paper);
    \item Visual results of IP2P and CLIPAway (referring to Sec \ref{sec:4-2} and Fig. \ref{fig:qualitative} in the main paper);
    \item Visual results on images of 512 × 512 resolution (referring to Sec \ref{sec:4-2} in the main paper);
    \item Failure cases (referring to Sec. \ref{sec:conclusion} in the main paper).
\end{itemize}

\section{Annotation page and user study page}
\label{sec:SP1}

\begin{figure*}[!h]
  \centering
   \includegraphics[width=\linewidth]{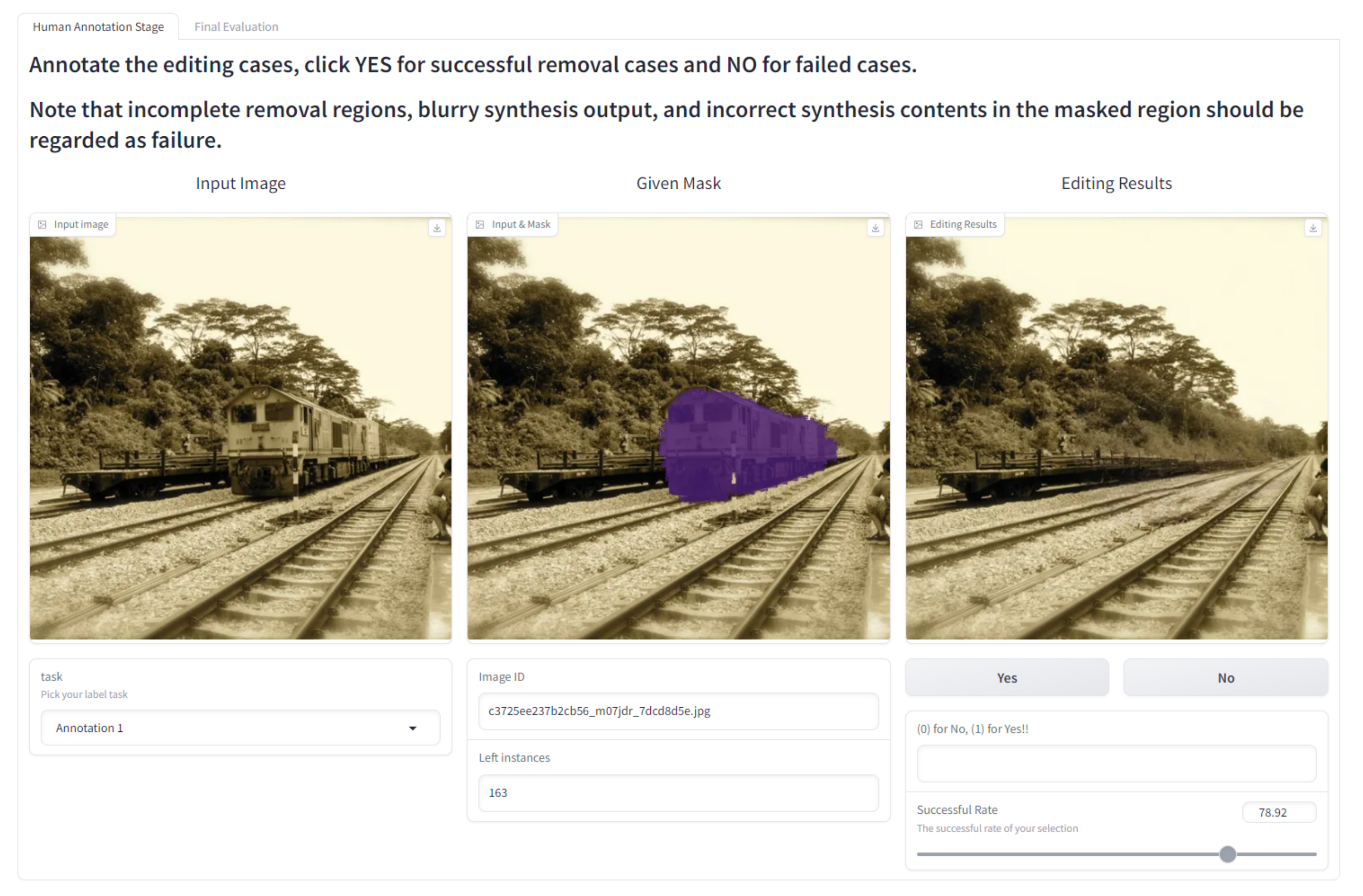}
\vspace{-3mm}
   \caption{The interface for human annotation. The annotators are asked to give feedback by clicking "Yes" or "No" button.}
   \label{fig:humanAnnotation}  
\end{figure*}

We design a webpage based on the open-source library Gradio \cite{abid2019gradio} to conduct the human annotation (referring to Sec. \ref{sec:human_annotation} in the main paper) and the final human evaluation (referring to Sec. \ref{sec:4-1} in the main paper). Annotators are provided with the original images, the mask images and the object removal results, as illustrated in \cref{fig:humanAnnotation}. They are asked to provide feedback by clicking the \textbf{Yes} or \textbf{No} button at the bottom right corner.

\begin{figure*}[!htbp]
  \centering
   \includegraphics[width=\linewidth]{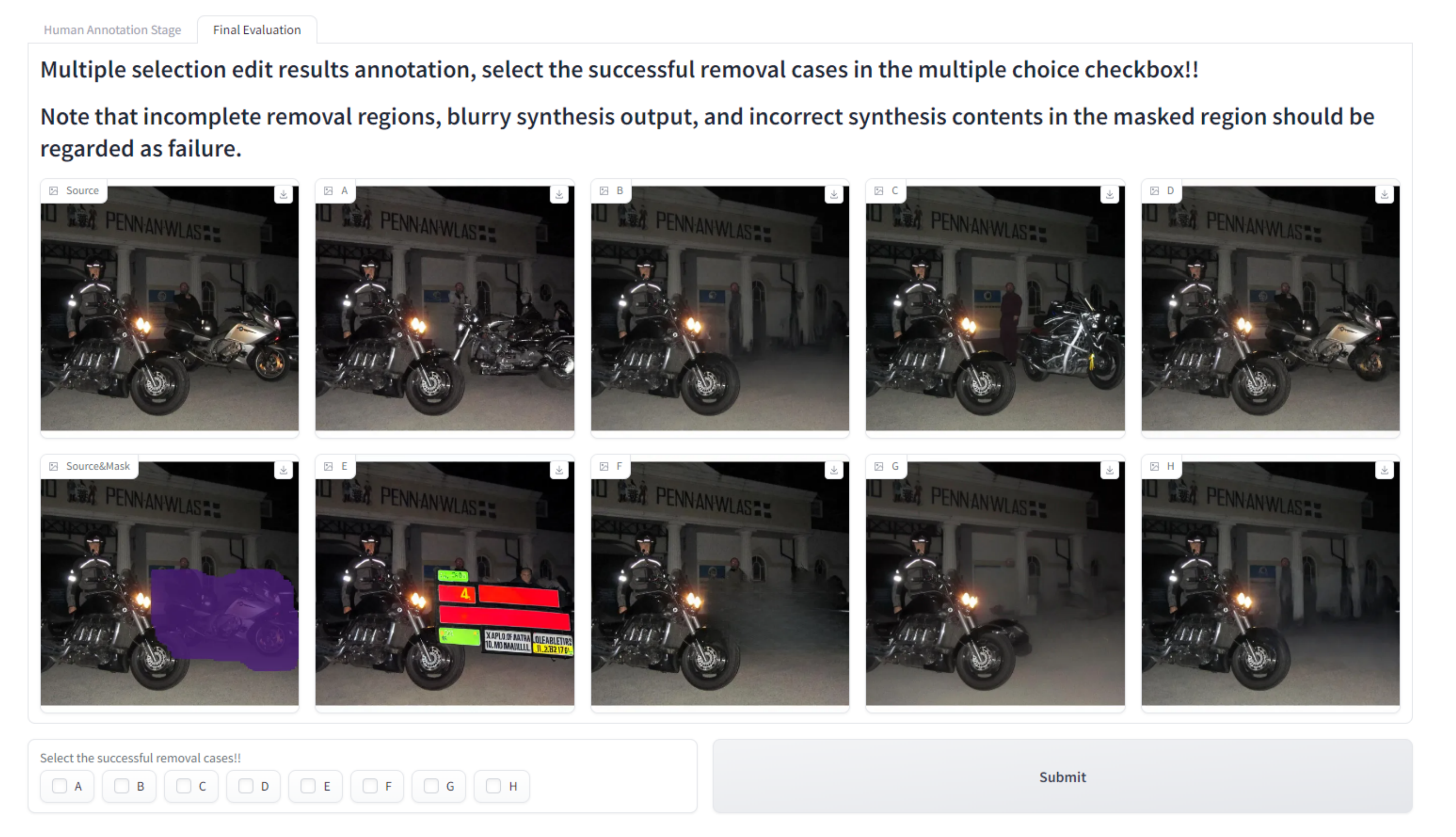}
\vspace{-3mm}
   \caption{The interface for human evaluation. The volunteers make selections by checking the multiple-choice check-boxes at the bottom left corner.}
   \label{fig:humanEvaluation}  
\end{figure*}

The interface for final human evaluation is shown in \cref{fig:humanEvaluation}. The input images and the masked images are displayed in the left column. The editing results of different methods as displayed in the right columns. Annotators are asked to click the multiple-choice check-boxes to select the successful removal results among different methods and submit the results. We randomly shuffle the display order in each evaluation. Five volunteers participated in the final evaluation, and each volunteer annotated 1,000 samples, including 500 pairs of object removal cases under $512 \times 512$ resolution and 500 pairs under $1024 \times 1024$ resolution. We calculate the average success rate for different methods based on these human evaluations.

\section{The evaluation metrics for the discriminator}
\label{sec:SP2}

\begin{figure*}[!htbp]
  \centering
   \includegraphics[width=0.62\linewidth]{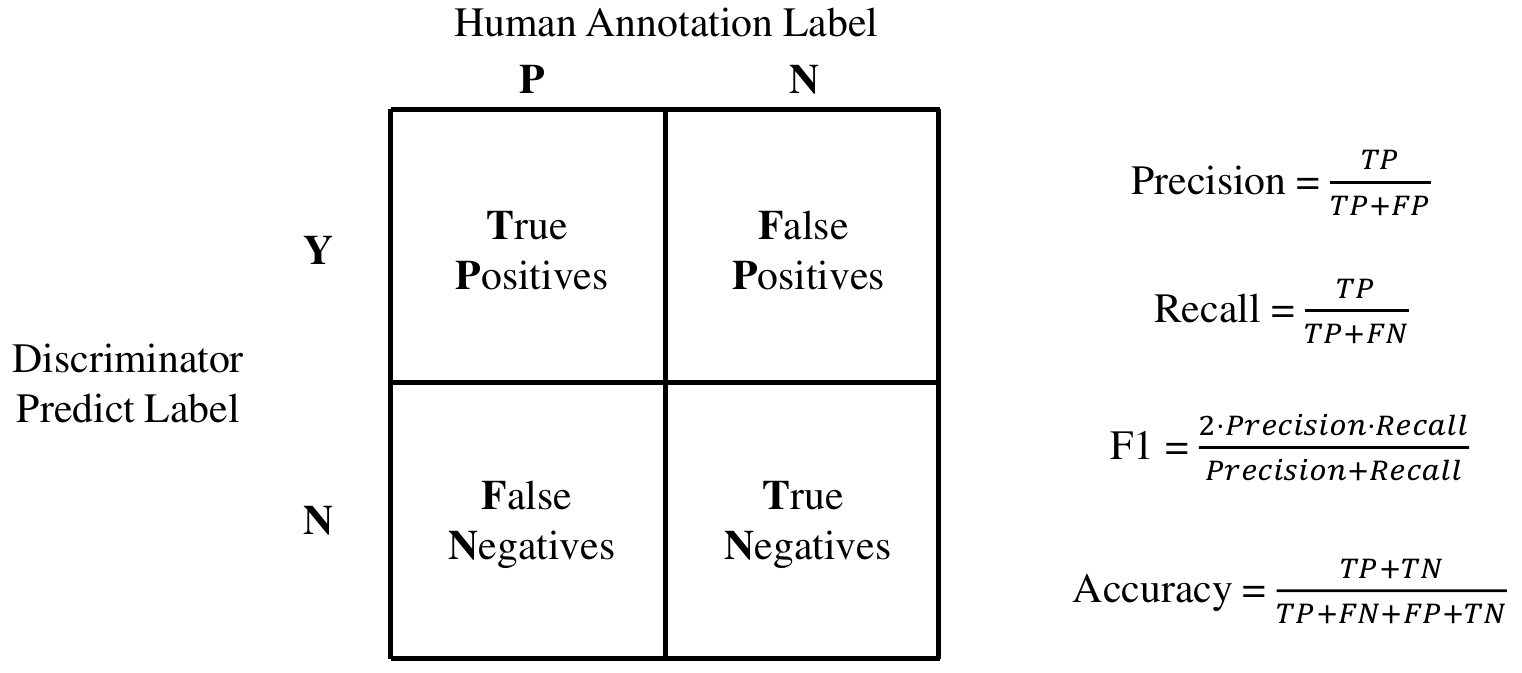}
   \caption{Confusion matrix and the definition of metrics for evaluating our discriminator. }
   \label{fig:evaluate_metrics}  
\end{figure*}

We use the 500 pairs in the test set with $512 \times 512$ resolution to test the discriminator. The edited results are generated by our RORem. The definitions of precision, recall, F1 and accuracy are illustrated in \cref{fig:evaluate_metrics}. Among these metrics, precision represents the percentage of the true positive samples to the total positive samples predicted by our discriminator. High precision ensures that the selected removal pairs are all of high-quality. By setting the threshold as 0.9, our final discriminator can reach a precision of 0.983, which allows us to obtain a large amount of high-quality data pairs. 

\section{Visual examples of object removal results at each training round}
\label{sec:SP3}

\begin{figure*}[thbp]
  \centering
   \includegraphics[width=\linewidth]{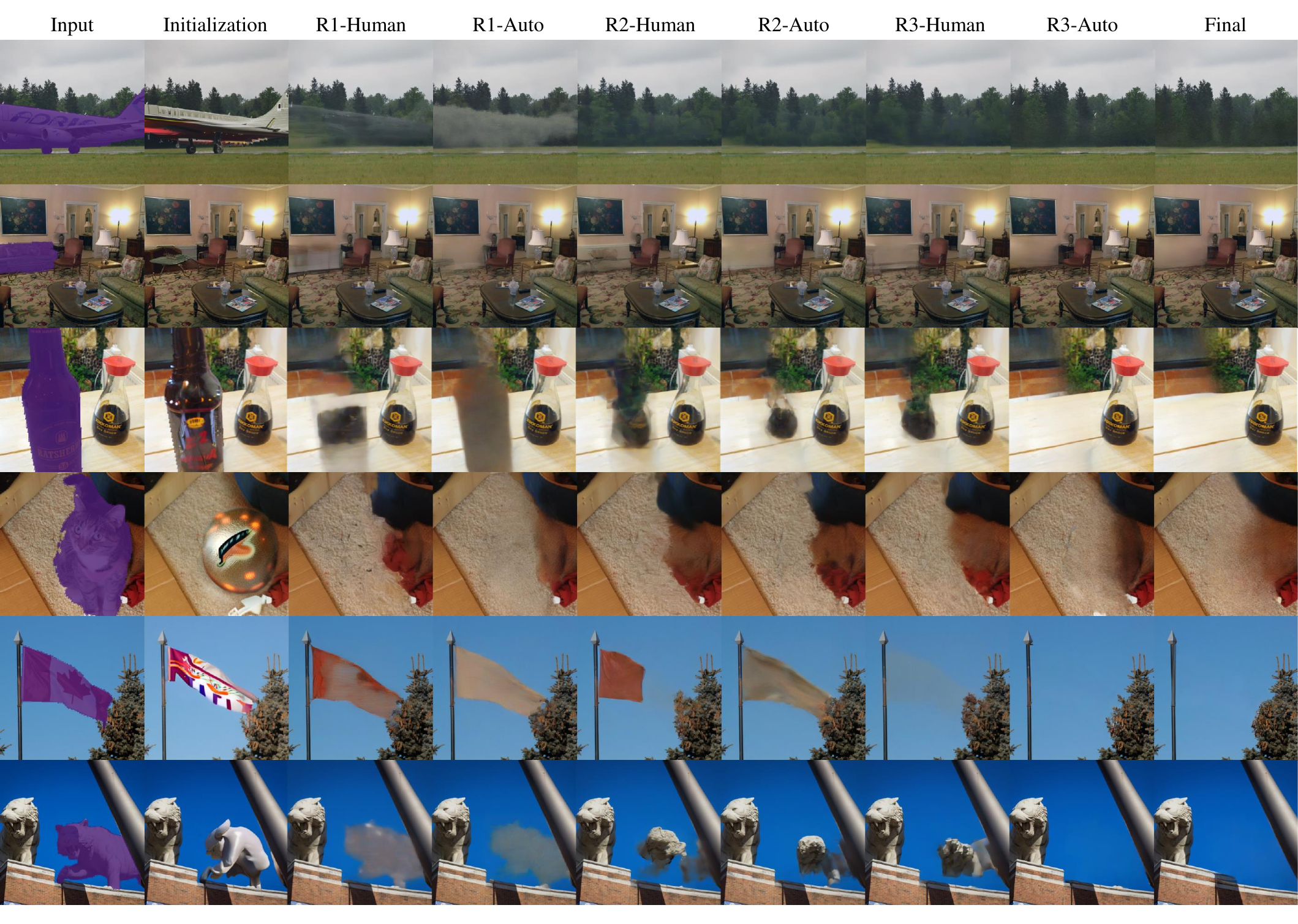}
   \caption{Visual results of RORem at each training round, one can see that the removal capacity of RORem improves with the increase of the dataset. }
   \label{fig:step_result}  
\end{figure*}

The visual examples of object removal results at each training round are provided in \cref{fig:step_result}. We can see that the initial model SDXL-inpainting \cite{podell2023sdxl} always fills the masked regions with semantically similar contents instead of removing it (column Initialization). After the first round of finetuning, RORem can successfully remove the selected sofa (row 2) and cat (row 4); however, its removal capacity is not good enough, leading to failures in other cases (see partial removal cases bottle, statue and blurry synthesis case airplane). After we  extend the training dataset and conduct more finetuning rounds (see column R1-Human to column R3-Auto), RORem can successfully remove the masked regions in most cases. Finally, we collect images with 2K resolution from DIV2K \cite{Agustsson_2017_CVPR_Workshops} and Flicker2K \cite{timofte2017ntire} to conduct the final stage finetuning, where the removal capacity of RORem can be well preserved (see column Final) and the image quality can be improved (see Tab. \ref{tab:train_data} in the main paper).

\section{Visual results of IP2P and CLIPAway}
\label{sec:SP4}

\begin{figure*}[thbp]
  \centering
   \includegraphics[width=\linewidth]{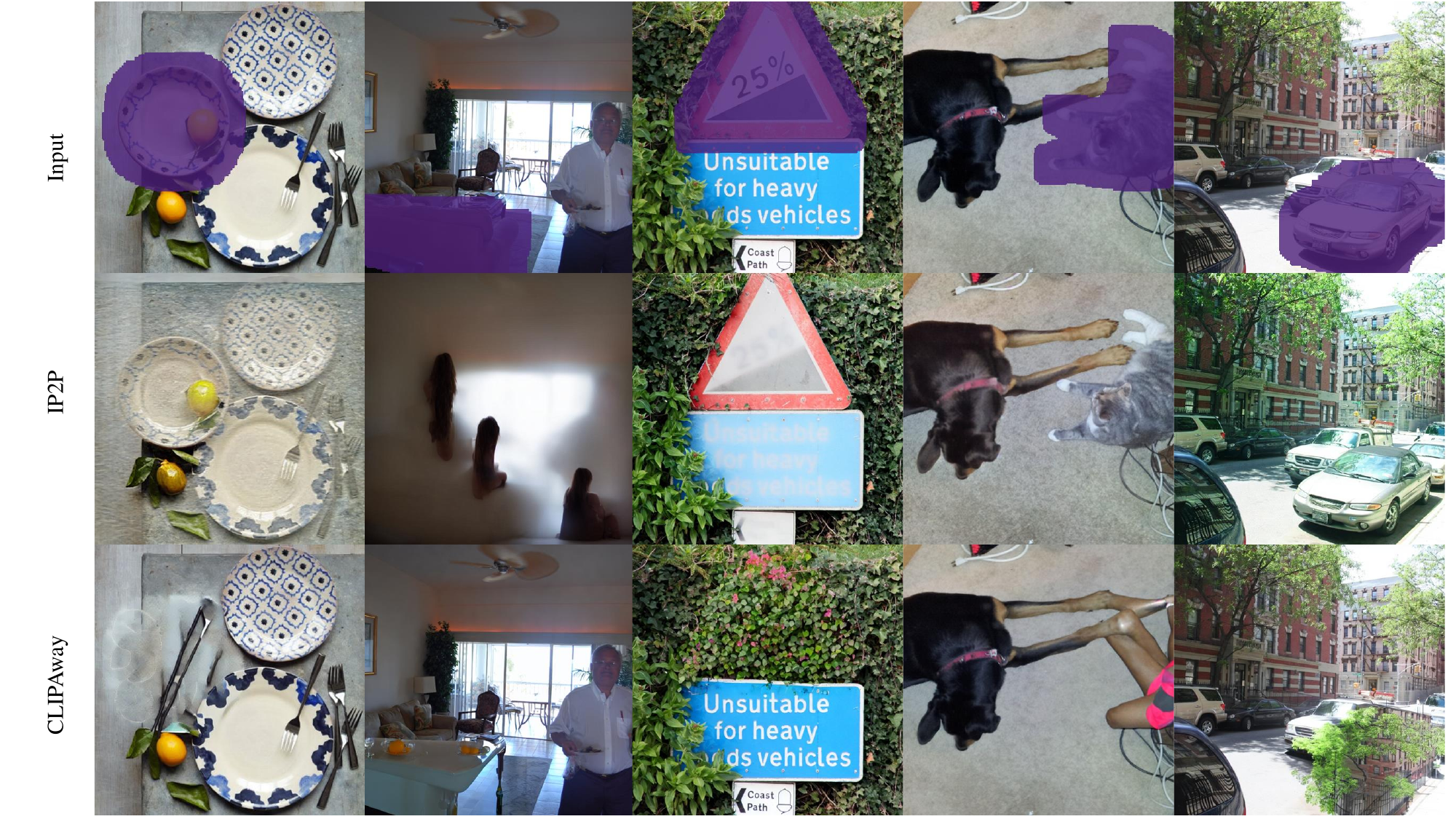}
   \caption{Visual results of IP2P and CLIPAway on $1024 \times 1024$ resolution images. }
   \label{fig:IP2P_CLIPAway}  
\end{figure*}

The visual editing results of IP2P and CLIPAway on images of resolution of $1024 \times 1024$ are illustrated in \cref{fig:IP2P_CLIPAway}. We can see that IP2P fails in all cases and even changes the overall style of the given images (see images plate in column 1 and car in column 5). CLIPAway exhibits the same problem as PPT, which often fills the masked regions with incorrect contents (see images sofa, dog and car).

\section{Visual results on images of $512 \times 512$ resolution}
\label{sec:SP5}

\begin{figure*}[thbp]
  \centering
   \includegraphics[width=\linewidth]{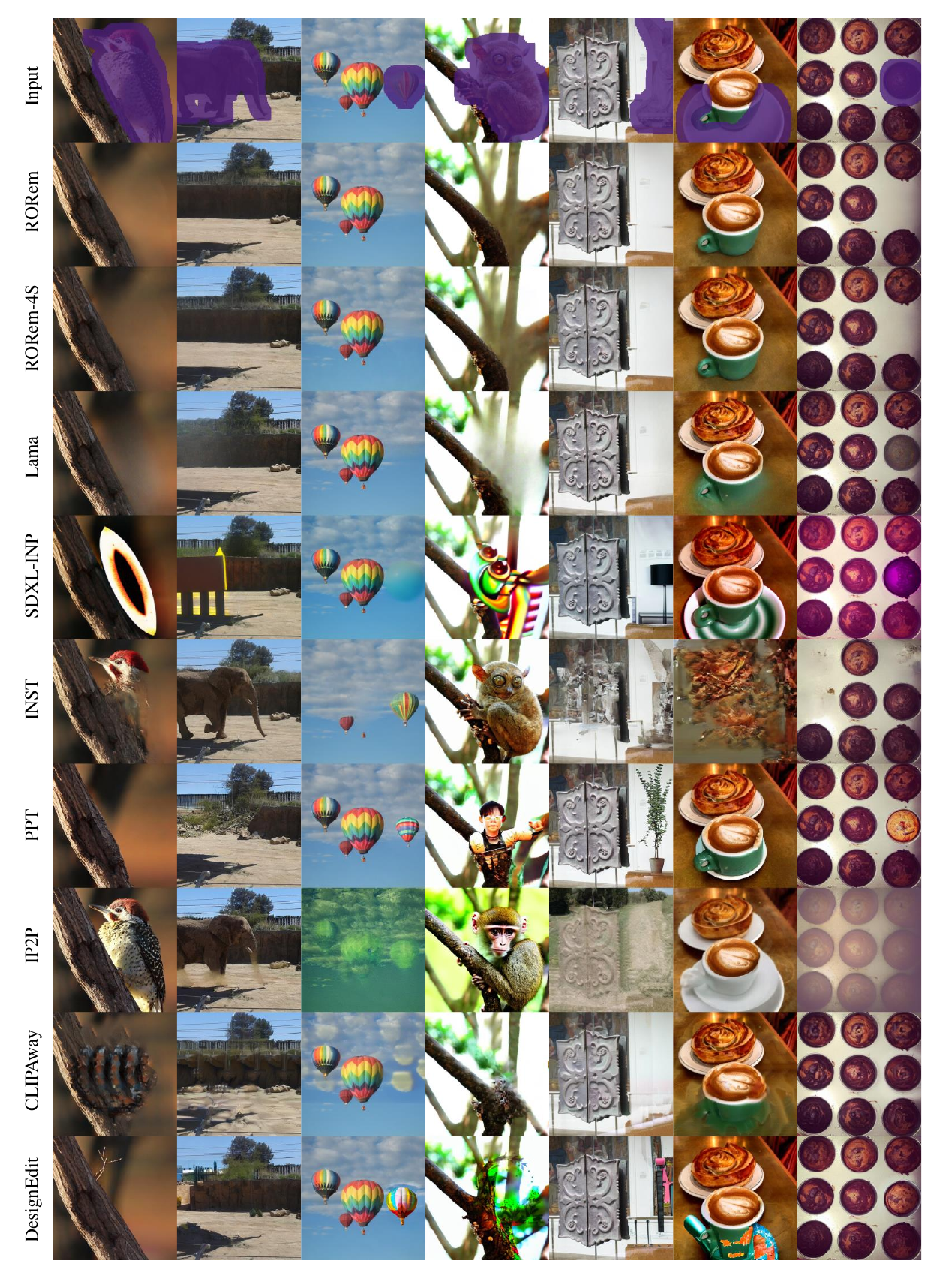}
   \vspace{-10mm}
   \caption{Visual comparison of the object removal results by RORem and other methods on $512 \times 512$ resolution images. One can see there can be incomplete removal regions, blurry synthesis output, wrong removal target, and incorrect synthese contents in previous methods, while RORem demonstrate robust removal performance.}
   \label{fig:visual_512}  
\end{figure*}

The qualitative comparisons on images of $512 \times 512$ resolution are illustrated in \cref{fig:visual_512}. We can see that Lama can generate blurry synthesis outputs in some cases (see images koala in column 4 and plate in column 6). SDXL-INP, IP2P and INST fail in most cases. Moreover, as INST and IP2P are text-driven removal methods, the ambiguity of text instructions can lead to removal failures of selected objects (see images hot air ballon in column 3 and cup cake in column 6). IP2P not only fails to remove the select objects but also changes the overall style and details of the original images (see images hot air ballon, koala, and cup cake). PPT and CLIPAway can fill the masked regions with nonexistent contents in images bird (column 1), koala (column 4) and statue (column 5). DesignEdit succeeds in the first two removal cases, however it suffers from visual artifacts (see images koala, plate). In contrast, RORem successfully removes the selected objects in most cases. Meanwhile, our distilled RORem-4S model also works well in these cases with less time overhead.

\section{Failure Cases}
\label{sec:SP6}

\begin{figure*}[thbp]
  \centering
   \includegraphics[width=\linewidth]{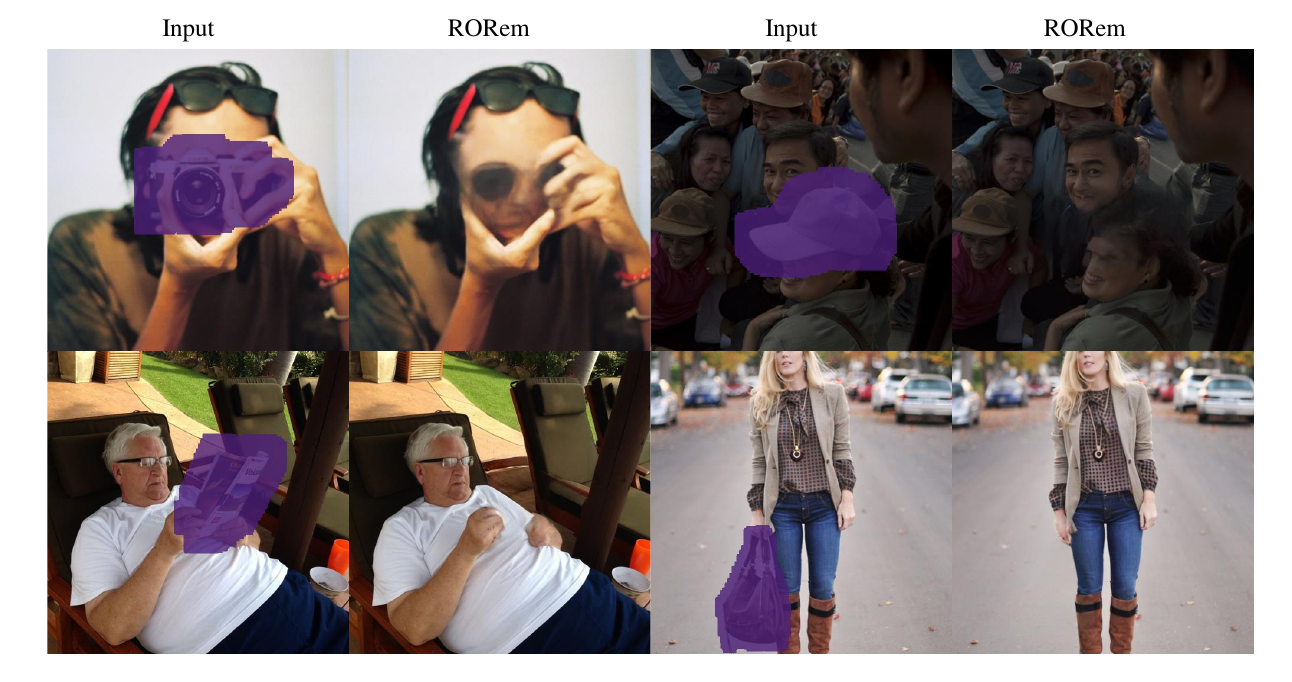}
   \caption{Failure cases of RORem.}
   \label{fig:visual_failure}  
\end{figure*}

As we stated in the conclusion section of the main paper, although RORem achieves great improvement on the overall removal performance, it may fail in cases when the background contains human fingers and faces. Some failure cases are depicted in \cref{fig:visual_failure}. Future work will be conducted for further improving the performance of RORem on these editing scenarios.

\clearpage

\end{document}